**Synthetic Iris Image Databases and Identity Leakage: Risks and Mitigation Strategies**

**Ada Sawilska, Mateusz Trokielewicz**

Warsaw University of Technology, Institute of Control and Computation Engineering, Nowowiejska 15/19, 00-665 Warsaw, Poland

PayEye Poland, Rondo ONZ 1, 00-124 Warsaw, Poland

**Abstract**

**This paper presents a comprehensive overview of iris image synthesis methods, which can alleviate the issues associated with gathering large, diverse datasets of biometric data from living individuals, which are considered pivotal for biometric methods development. These methods for synthesizing iris data range from traditional, hand crafted image processing-based techniques, through various iterations of GAN-based image generators, variational autoencoders (VAEs), as well as diffusion models. The potential and fidelity in iris image generation of each method is discussed and examples of inferred predictions are provided. Furthermore, the risks of individual biometric features leakage from the training sets are considered, together with possible strategies for preventing them, which have to be implemented should these generative methods be considered a valid replacement of real-world biometric datasets.**

**Keywords: biometrics, iris recognition, generative AI, iris image synthesis, biometric datasets**

1. **Introduction**

Imagine a world where artificial identities exist, crafted with such precision that they are indistinguishable from real ones. From deepfake videos to AI-generated faces, the ability to synthesize human-like features has reached an unprecedented level of sophistication. But what happens when this technology extends to something as personal and unique as our irises? We are already there.

Iris recognition has long been heralded as one of the most secure biometric authentication methods [1]. Unlike fingerprints, whose traces remain on everything we touch, making them vulnerable, or facial recognition, which can be affected by lighting variations, facial expressions, aging, and occlusions (*e.g.,* masks or glasses) [2], the intricate patterns of the human iris remain remarkably stable throughout life. This makes iris-based identification highly secure and reliable. However, deep learning-based biometric systems need enormous amounts of samples from real individuals during the training phase to perform accurately. Collecting such data is not only organizationally complex, requiring informed consent and strict privacy measures, but also legally challenging, as biometric data is classified as sensitive information under the General Data Protection Regulation (GDPR) in the European Union [3]. The GDPR imposes stringent requirements for the collection, processing, and storage of biometric data, including explicit consent and strict security measures to prevent unauthorized access. Additionally, processing biometric data must meet specific legal justifications, making large-scale data acquisition difficult and time consuming. As artificial neural networks research advances, scientists have started generating synthetic iris datasets to train and improve recognition systems while addressing privacy concerns.

The promise of synthetic iris datasets is immense: they offer a way to build robust biometric models without relying on real-world samples that might compromise one's privacy [4]. This innovation, however, comes with its own set of risks. If synthetic images retain identifiable features from real individuals they have been trained on, identity leakage can occur, undermining the privacy these systems aim to protect in the first place. In this paper, we delve into the process of synthetic iris generation, exploring how these datasets are created, the challenges they present, and the solutions that can safeguard our identities in an era where the line between real and artificial is increasingly blurred.

## 2. Historical background

John Daugman's work from 1993 on iris recognition was a breakthrough and became the foundation for modern iris biometrics [1]. Initial iris recognition systems relied solely on real iris databases collected from subjects but in 2003 the first attempt to create synthetic iris was made. Lefohn *et al.* based their knowledge on the procedure developed by artificial eye makers - ocularists - for physical iris synthesis that results in artificial eyes with all the crucial appearance characteristics [5]. They introduced a 3D computer model of the human iris using stacked conical surfaces with 30–70 layers of painted semi transparent textures, which are scanned, converted into opacity maps, and rendered with ray tracing for realistic depth and appearance, see Fig. 1. The goal was not yet to deliver synthetic iris databases but rather to easily create realistic looking irises for both the ocular prosthetics and entertainment industries.

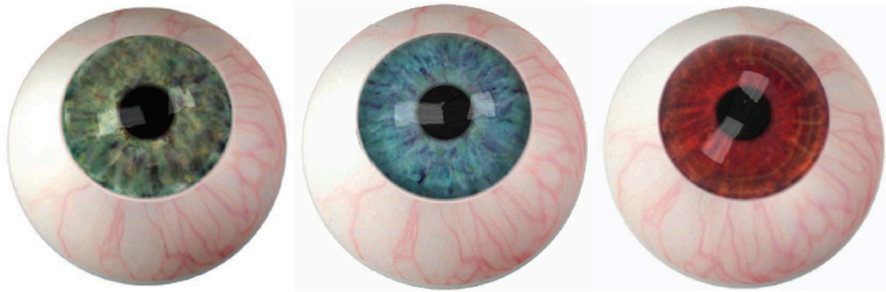

Fig. 1. Visualizations of artificial eyes generated using a 3D layered modeling technique, illustrating natural iris color variations [5].

In 2004 the idea of creating a synthetic iris database occurred in the literature for the first time. Cui *et al.* introduced the method to synthesize iris images using Principal Component Analysis (PCA) to construct coarse iris patterns by controlling coefficients in a high-dimensional space and then improving the image with super-resolution [6]. Makthal and Ross introduced a method utilizing the Markov Random Field (MRF) model to capture the complex texture of the iris, employing textural patterns to generate synthetic iris images from an initial random noise distribution [7]. Shah and Ross refined the previous method by employing the MRF model to generate a background texture that captures the global appearance of the iris. Key anatomical features, including radial and concentric furrows, the collarette, and crypts were then incorporated into the synthetic images. Additionally, line integral convolution was applied to enhance the texture of the radial furrows, ensuring a more realistic representation of iris patterns [8].

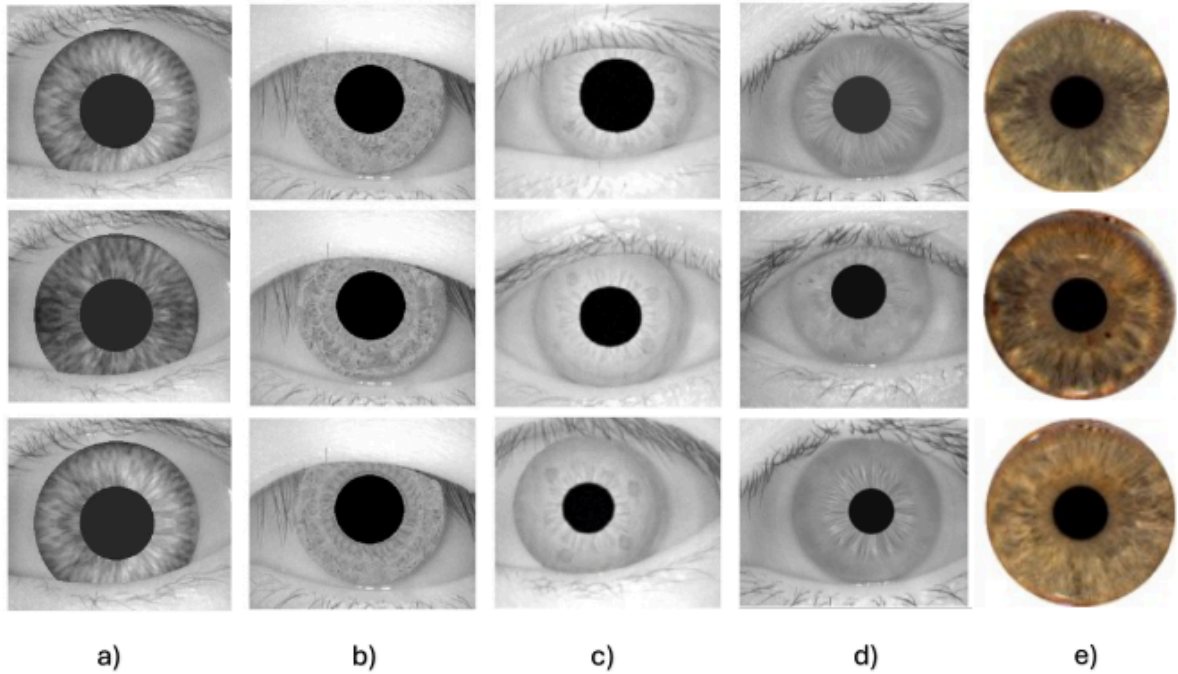

Fig. 2. Synthetic irises generated with methods based on a) PCA [6], b) MRF [7], c) MRF with feature agglomeration [8], d) anatomical-based approach [9, 10], e) reverse subdivision [11, 12].

Zuo and Schmid based their method on the anatomical approach where they synthesized the eye image by modeling the iris structure with continuous fibers in cylindrical coordinates, projecting them into a 2D space, and shaping the pupil and the iris. They further refined the image by adding surface details such as the collarette, a semi-transparent top layer, and randomly generated eyelids and eyelashes. The resulting synthetic irises were evaluated using Independent Component Analysis (ICA) and a Gabor filter-based recognition system, demonstrating their similarity to real iris images [9, 10]. Wecker *et al.* proposed a method for generating synthetic iris images using a multiresolution technique called reverse subdivision. This approach decomposes real iris images into multiple levels of detail and then recombines different components from various irises to create new, unique iris images [11, 12], see Fig. 2. Wei *et al.* used a patch-based sampling approach, where small patches from real iris images are combined to create a synthetic iris [13]. Intra-class variations of newly created iris are introduced through transformations such as deformation, defocus, noise addition, and rotation, ensuring diversity in the generated dataset, see Fig. 3.

All the proposed approaches aim to synthesize iris textures that closely resemble real-world patterns to augment existing iris databases. While some methods generate textures from scratch, others leverage patches from existing iris images as a basis for constructing novel and unique iris textures.

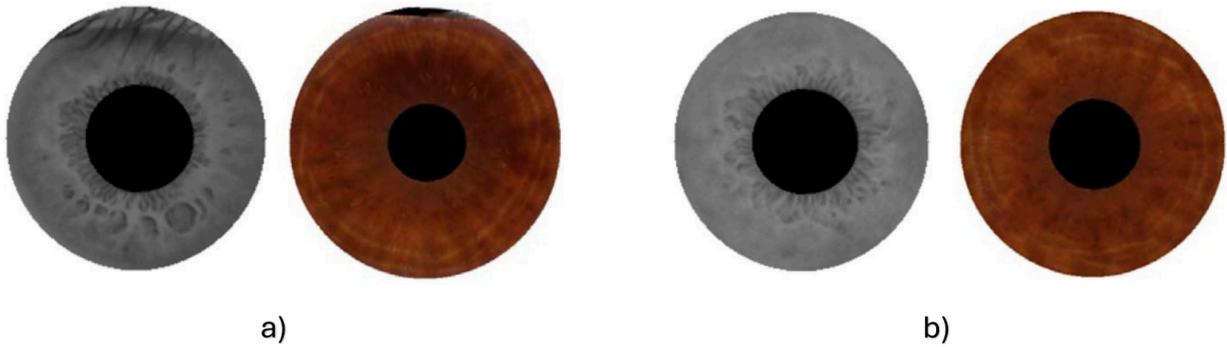

Fig. 3. The images a) are real iris images, while b) are synthetic irises generated through patch-based sampling of real iris textures [13].

Another goal of generating synthetic iris textures can be to create an artificial iris that can deceive a recognition system. This topic was covered in the articles by Venugopalan *et al*. [14] and Galbally *et al*. [15].

In 2011 Venugopalan *et al*. started synthesizing a spoofed iris from the iris feature space, represented by the iris bit code. The goal is to reverse-engineer an image, using the prior knowledge about the recognition system, so that, when processed with the same feature extraction method, produces a similar iris code. Since the iris code contains only phase information, the missing magnitude is obtained from any other given iris texture. By controlling the amount of texture embedded, the discriminating pattern can be hidden until it is visually indiscernible from the background [14], Fig. 4. Galbally *et al*. proposed a genetic algorithm-based probabilistic approach to reconstruct synthetic iris images from binary iris codes where no a priori knowledge is needed [15]. The method considered iris reconstruction as an optimization problem, where the objective was to generate an iris image whose extracted iris code closely matched a given real iris code. The genetic algorithm iteratively improved candidate synthetic images by applying selection, crossover, and mutation operations, optimizing the similarity score between the real and synthetic iris codes until a predefined threshold was met, see Fig. 4.

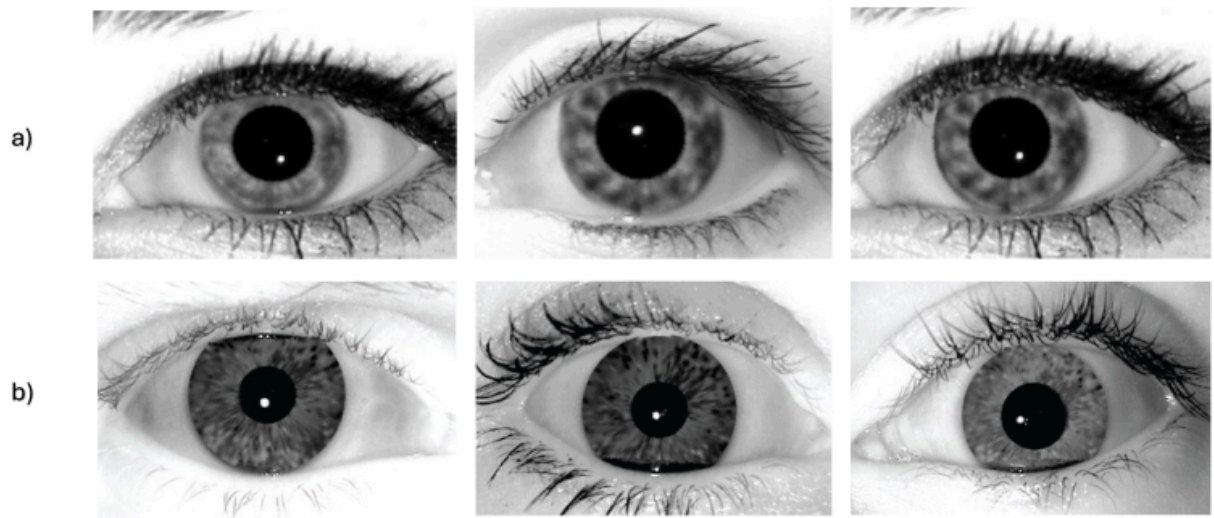

Fig. 4. Iris images reconstructed from binary iris codes. Synthetic irises generated by embedding external texture to complement phase information [14] (top row), reconstructions using a genetic algorithm to optimize the match with genuine iris codes [15] (bottom row).

Proenca and Neves provided a method of generating synthetic iris codes based on bit correlations and demonstrated that the created samples attain some of the desired statistical properties, such as the shapes of genuine and impostor distributions [16]. Drozdowski *et al*. proposed the SIC-Gen method which generates synthetic iris codes by first creating a base template with alternating bit sequences, then iteratively flipping bits to introduce realistic variations while maintaining statistical properties such as bit correlation, degrees of freedom, and Hamming distance distributions [17]. Target Hamming distances guide the process, ensuring controlled similarity between mated templates, followed by smoothing and post-processing to add noise, occlusions, and shifts. This approach produces synthetic iris codes that closely resemble real biometric data in both structure and statistical distribution. Real irises are so anatomically complex patterns and present a significant challenge for mathematical representation, making it nearly impossible to precisely describe any individual pattern using traditional methods [10]. This complexity called for an innovative approach—one that could capture the underlying distribution of real-world iris textures rather than relying on rigid mathematical formulas.

## 3. Methods utilizing generative AI

A major breakthrough in generative modeling arrived in 2014 with the introduction of generative adversarial networks (GANs) by Goodfellow *et al.* [18]. This framework establishes a dynamic competition between two deep neural networks: a generator, which strives to produce increasingly realistic data sampling from a given distribution, and a discriminator, which attempts to distinguish between genuine and artificially generated samples. Through this continuous interplay, the generator progressively refines its output until it becomes indistinguishable from real data, effectively learning to replicate complex patterns such as those found in natural images. Yadav and Ross conducted a comparative analysis of all GAN-based models [4]. Their study systematically evaluates a variety of GAN architectures in terms of image realism, biometric uniqueness, and utility in iris recognition and presentation attack detection. They demonstrate that while several models produce visually convincing outputs, only a few, such as iWarpGAN, successfully balance image quality with identity diversity, making them especially valuable for biometric applications. Alongside GANs, advancements in variational autoencoders (VAEs) further propelled the field of generative modeling, offering powerful new tools for creating realistic images and textures [19]. Autoencoders work by compressing input data into a lower-dimensional latent space (encoding) and then reconstructing it back to the original form (decoding), learning an efficient internal representation in the process. Another generative method which has gained attention for its ability to generate high-quality images by gradually transforming random noise into clear, detailed data are diffusion models [20, 21]. They work by learning how to reverse an iterative noising process, allowing them to produce realistic and diverse results. These techniques revolutionized the ability to synthesize high-quality visual data, opening doors to applications that had previously been beyond reach, from generating lifelike human faces [22, 23] to reconstructing intricate biological structures [24, 25].

The initial attempts at synthetic iris image generation utilized Deep Convolutional GANs (DCGANs) [24, 26]. Kohli *et al.* evaluated the effectiveness of the proposed iDCGAN (Iris Deep Convolutional Generative Adversarial Network) for generating synthetic iris images and their impact as a presentation attack on iris recognition systems [26]. The model was trained using a dataset of 8,905 real iris images and generated an equal number of synthetic iris images. The study demonstrates that iDCGAN-generated synthetic iris images closely resemble real irises, making them highly effective for biometric attacks. The quality metrics analysis, based on ISO/IEC 29794-6 standards [27], shows a strong overlap between the distribution of real and synthetic images, with χ² distances as low as 0.34 (pupil circularity) and 1.07 (sharpness, pupil-to-iris ratio), indicating high visual similarity. When tested on VeriEye, a popular commercially available matcher [28], synthetic images scored a 67.66% false acceptance rate, meaning they could successfully bypass the recognition system. Additionally, the state-of-the-art DESIST attack detection system struggled to distinguish them, with an equal error rate (EER) of 14.19%, almost twice that of previous synthetic datasets (7.09%) [29]. These results highlight the high realism and security risk posed by iDCGAN-generated iris images, emphasizing the urgent need for improved attack detection mechanisms in biometric security.

Minaee *et al.* in the Iris-GAN study presents a deep convolutional GAN (DC-GAN) framework for generating highly realistic synthetic iris images [24]. The model was trained on two datasets: CASIA-Iris (20,000 images) and IIT Delhi Iris (1,120 images), producing visually convincing iris images over 120 and 140 training epochs, respectively. The Frechet Inception Distance (FID) scores - a metric assessing generated image quality and diversity by comparing their statistical distributions to real images [30] - were 42.1 for CASIA and 41.08 for IIT Delhi, indicating that the generated images are somewhat realistic but not perfect. With each epoch the model improved in pupil shape, eyelash details, and texture realism highlighting the potential of synthetic iris images and their possible use in biometric spoofing or dataset augmentation.

Yadav *et al.* utilized Relativistic Average Standard Generative Adversarial Network (RaSGAN) for generating synthetic iris images and evaluated their impact on presentation attack detection (PAD) systems [31]. Unlike traditional GANs, RaSGAN improves the network's ability to generate realistic data by using a relativistic discriminator. Instead of simply determining whether a sample is real or fake, this discriminator compares real and synthetic data, trying to maximize the probability that real samples appear more realistic than fake ones and vice versa. The model was trained on 2,778 authentic iris images from the Berc-iris-fake dataset [32] and produced 6,277 synthetic samples, achieving a FID score of 39.17, indicating a closer resemblance to real data. Experimental results show that PAD algorithms like DESIST, BSIF+SVM, Iris-TLPAD and pre-trained VGG-16 struggle to distinguish these synthetic images from real ones. When synthetic iris images were used as previously unseen presentation attack scenario, EERs reached 50.64% for BSIF and 57.45% for DESIST, revealing their strong potential as threat vectors. The study concludes that RaSGAN-generated irises can deceive biometric systems and can be leveraged to improve PAD generalization against unseen attacks.

Lee *et al.* employed Conditional Generative Adversarial Networks (cGANs) to augment iris datasets by generating identity-preserving synthetic images with controlled geometric variation [33]. Unlike standard GANs, cGAN was conditioned on normalized iris images in polar coordinates, combined with systematically shifted iris and pupil center positions (±4 pixels) applied during the normalization process, to simulate natural eye movement. The model was trained on 6676 real iris images and each shifted input was paired with the corresponding identity's geometric center image - an average representative sample, used as the ground truth to guide the generator toward producing realistic, undistorted iris patterns. This conditioning helped guide the image generation process, ensuring that the synthetic iris images maintained structural consistency with real iris patterns while introducing controlled variations. Once trained, the generator was used to synthesize new, realistic normalized iris images that preserved identity while introducing controlled variations, effectively expanding the training dataset for iris recognition. Experiments across NICE.II, MICHE and CASIA-Iris-Distance datasets demonstrated that synthetic images improved recognition performance beyond traditional augmentation methods such as geometric transformations and brightness adjustments, which are limited in capturing the complex intra-class variability of iris patterns. The Equal Error Rate (EER) was reduced by 1.23%, 1.57%, and 1.51% on each dataset respectively, indicating an improvement in recognition performance. Additionally, the d-prime value increased, reflecting a better separation between genuine and impostor matches, rising from 2.89 to 3.14 for NICE.II, 2.54 to 2.79 for MICHE, and 2.19 to 2.45 for CASIA-Iris-Distance.

Li *et al.* explored the use of Conditional Wasserstein Generative Adversarial Networks with Gradient Penalty (CWGAN-GP) to address class imbalance in iris image datasets by generating synthetic images of rare cases, such as eyes with glasses or unusual pupil sizes [34]. It extends Wasserstein GAN with Gradient Penalty (WGAN-GP) [35] by conditioning both the generator and discriminator on class labels. WGAN-GP is an improved GAN that stabilizes training by using Wasserstein distance with a gradient penalty, preventing vanishing/exploding gradients and enhancing sample quality [35]. The obtained results show that CWGAN-GP effectively generates high-quality images that closely resemble real iris images, improving dataset diversity and enabling better training for deep learning models. The proposed method achieved an FID score of 0.690, significantly lower than previous methods (*e.g.,* 42.1 in prior works [24]), indicating superior image quality. The authors claim that CWGAN-GP can generate an unlimited number of realistic iris images, facilitating robust iris recognition models and improving deep learning performance in biometric applications.

Yadav *et al*. proposed a Cyclic Image Translation Generative Adversarial Network (CIT-GAN) which is a novel multi-domain style transfer GAN designed to generate high-quality synthetic images for iris presentation attack detection (PAD) [36]. CIT-GAN improves the basic GAN by making it better suited for translating images from one category to another. It introduces a Styling Network that learns domain-specific characteristics, allowing the Generator to modify existing images instead of generating them from noise. Additionally, it includes a multi-branch Discriminator for domain-aware classification and cycle consistency loss to ensure transformed images retain key features, making it highly effective for multi-domain image translation in iris presentation attack detection. The generated samples from CIT-GAN obtained an average FID score of 32.79.

Wang *et al*. presented an iris image generation method using StyleGAN2 with contrastive learning, enabling the synthesis of realistic intra- and inter-class iris images while preserving identity features [37]. A dual-channel input protocol separates iris topology and texture, improving image quality and variation control. The model achieves an FID score of 5.27, significantly better than previous method which obtained FID equal 39.17 [31], proving higher realism. The study confirms that synthetic iris datasets can enhance recognition accuracy, making them a valuable alternative to expensive real data collection.

Maureira *et al*. performed an analysis of the most used GAN architectures and stated that the best is StyleGAN2 [38].

Khan *et al.* developed DeformIrisNet which is a U-Net-based autoencoder trained to realistically deform iris textures due to pupil size variations, without relying on anatomical assumptions [39]. It takes an original iris image and a target pupil shape mask and applies learned non-linear deformations while ensuring that biometric identity is preserved. DeformIrisNet is trained using Warsaw-BioBase-Pupil-Dynamics v3.0 [40, 41, 42], a dataset containing over 117000 high-quality grayscale iris images with pupil sizes varying from 0.2 to 0.7 pupil-to-iris ratio. To generate realistic and sharp iris images the model is trained with identity-preserving loss functions and perceptual constraints like Learned Perceptual Image Patch Similarity (LPIPS) which ensure the generated image has a similar high-level appearance to the original iris, and mask consistency loss ensures that the deformed iris matches the target mask shape. The performance of DeformIrisNet was compared against two baseline normalization methods: the standard linear rubber-sheet model [1] and a biomechanical deformation model [43]. The comparison focused on recognition accuracy using binary iris codes. The proposed method achieved a lower EER of 11.8%, compared to 12.1% for the linear model, but the biomechanical model gained a lower score of 11.5%. Additionally, DeformIrisNet gave the highest decidability score ($d' = 3.003$), indicating better separation between genuine and impostor comparisons and improved identity preservation under varying pupil conditions.

Yadav and Ross introduced a novel method for generating realistic and diverse synthetic iris images by separating identity and style features [44]. Traditional GANs often struggle with generating diverse and unique iris identities, frequently producing images that closely resemble those in the training dataset [45]. To address these limitations, the authors proposed iWarpGAN, which employs two distinct transformation pathways which are later combined in the generator, namely:

- Identity Transformation, which is designed to generate unique iris identities that are not present in the training set. It achieves this by learning a Radial Basis Function (RBF)-based warp function in the latent space of a GAN, enabling the creation of new identities,
- Style Transformation that extracts the style code from a reference iris image, capturing attributes such as texture and color.

The results demonstrated that iWarpGAN effectively generates high-quality, unique identities with intra-class variations which do not resemble real, training set irises.

Anderson *et al.* introduced a specialized Identity-Preserving GAN architecture aimed at domain-translation, specifically between visible and near-infrared (NIR) images, keeping characteristics of identity intact [46]. Their method was evaluated across three cross-spectral iris datasets: PolyU, Cross-Eyed, and the WVU Multispectral Iris dataset, demonstrating improvements in cross-spectral recognition performance dropping the EER from 23.31% to 5.50% for PolyU data when applying score fusion with identity-preserving translation. On the Cross-Eyed dataset, the EER decreased from 9.95% to 2.50%, and for WVU dataset, despite being more challenging, EER reduced from 27.33% to 15.87% for normalized iris images, showcasing the potential of the model in real-world data conditions. These results affirm that the classifier component significantly boosts the generator's ability to retain biometric identity during spectral translation.

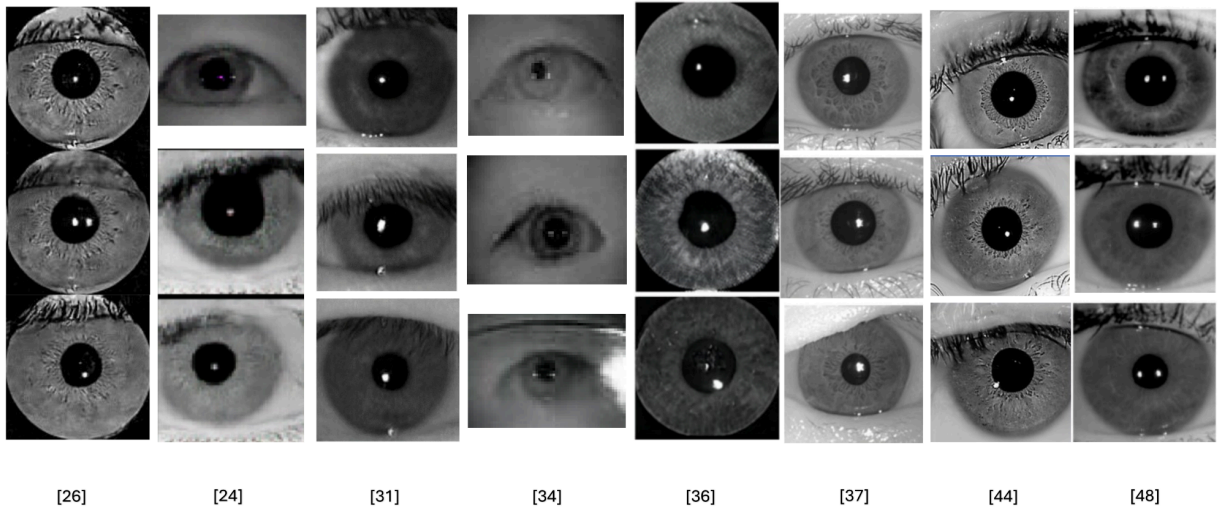

Fig. 5. Comparison of chosen synthetic irises generated with GANs. Source: [26, 24, 31, 34, 36, 37, 44, 48]

Kordas *et al.* explored the generation of synthetic iris images using the StyleGAN3 model, employing both Cartesian and polar representations [47]. This dual approach is novel, as previous research has primarily concentrated on Cartesian representations. A dataset of 11,623 iris images was used for training. Instead of using a fixed γ (gamma) regularization parameter, the researchers introduced a "gamma gradient" method, allowing dynamic adaptation where gamma was updated in step increments across different training stages. This approach helped maintain a balance between image sharpness and texture variation, avoiding excessive blurring or unrealistic features. The polar representation (FID = 8.75) showed a better match to real iris data than Cartesian (FID = 18.37). The distribution of similarity scores showed that synthetic polar iris images were visually close to real images but did not replicate identities.

Khan *et al.* introduced a hybrid deep learning model that combines an autoencoder-based deformation network with a GAN-based synthesis model to generate identity-preserving iris images while accounting for non-linear deformations caused by changes in pupil size [48]. The autoencoder learns to realistically deform iris textures when pupil size changes, using a triplet-based loss and perceptual constraints to ensure realism and biometric consistency. Meanwhile, the GAN component (StyleGAN2-ADA and StyleGAN3) synthesizes entirely new irises while preserving identity, making EyePreserve the first model to both generate and deform iris images with high biometric fidelity - that is, the ability to maintain consistent biometric identity across variations in pupil size, confirmed with iris matchers such as Human-Driven Binary Image Feature extractor (HDBIF) [49] and Dynamic Graph

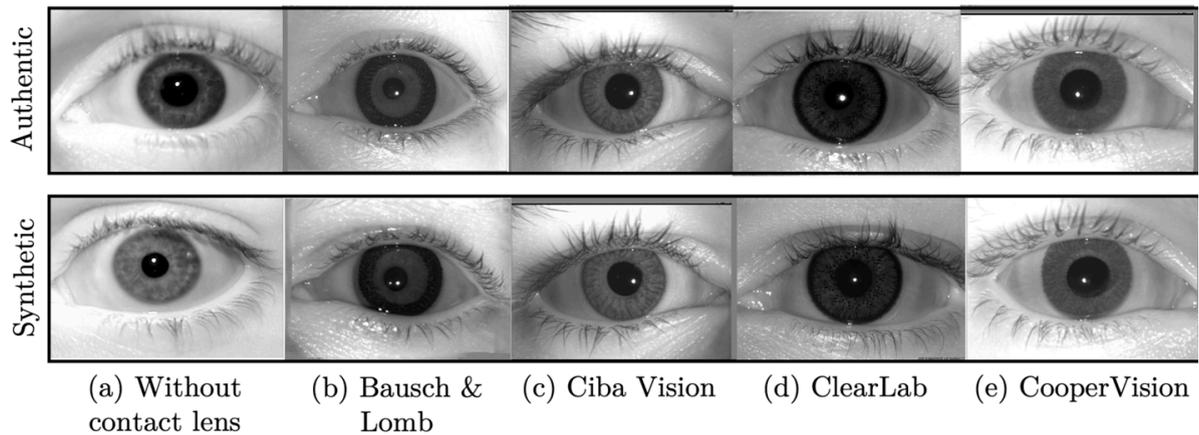

Fig. 6. Examples of authentic and synthetically-generated by a conditional StyleGAN2-ADA samples without textured contact lens and with textured contact lens of a given brand [51].

Representation for Occlusion Handling in Biometrics (DGR) [50], which showed strong genuine-impostor separation, high similarity scores for same-identity pairs, and improved decidability (d′) and AUC-ROC values compared to linear and biomechanical deformation baselines. This approach eliminates the need for prior anatomical assumptions and instead learns from real iris data. The comparison of the generation performance of the described GAN methods is illustrated in Fig. 5.

Mitcheff *et al*. introduced an approach to iris Presentation Attack Detection (PAD) that prioritizes privacy by entirely removing the need for authentic iris images [51]. The authors proposed a fully synthetic, privacy-safe iris PAD method trained solely on synthetic, identity leakage-free iris images generated using StyleGAN2-ADA models. They utilized two generative models, one to synthesize realistic iris images without textured contact lenses (bona fide samples) and another conditional model to produce iris images wearing textured contact lenses from seven known lens brands, Fig. 6. The incorporated method for identity-leakage mitigation systematically removed any synthetic images that biometrically resembled the authentic samples originally used to train the generative models. Neurotechnology's VeriEye SDK [52] was used for this purpose. The synthetic PAD models were built using three popular neural network architectures: DenseNet, ResNet, and Vision Transformer (ViT), and were trained on a set of 8,334 synthetic images. The experimental results demonstrated that PAD classifiers trained solely on synthetic images achieved an Area Under the ROC Curve (AUC-ROC) between 0.90 and 0.93. Although lower than models trained on an equivalent number of authentic iris samples, which reached an AUC of 0.97, this performance gap suggests the potential of using fully synthetic datasets for iris PAD training.

Bhuiyan and Czajka developed a conditional StyleGAN2-ADA-based synthesis model specifically designed for post-mortem iris recognition [53]. Their goal was to facilitate progress in this area by addressing the issue of data scarcity, simplifying the generation of realistic, forensic-alike iris samples. The training dataset combined three publicly available post-mortem iris datasets [54, 55, 56], totaling 8064 NIR iris images from 338 deceased individuals, categorized into 18 different Post-Mortem Interval (PMI) ranges from 0 to 1650 hours after death. To generate variations of iris images belonging to the same identity, the authors modified the StyleGAN latent space by introducing small perturbations ($\varepsilon$) within a defined hypersphere radius around the original latent vectors. These perturbations allowed the model to produce multiple realistic iris images of the same identity with subtle differences, closely mimicking within-class variations found in authentic forensic iris samples. The optimal radius ($\varepsilon\_max$) of these perturbations was experimentally determined to align the distribution of synthetic images with authentic data, ensuring realistic and identity-consistent results. The model produced a dataset of 180,000 synthetic post-mortem iris images across 18 PMI intervals, which contained 10,000 images from 1,000 identities.

Li *et al*. addressed the problem of iris image partial occlusion due to reflections, eyeglasses or poor lighting, which degrades the performance of iris recognition systems [57]. Traditional methods typically ignore these regions but do not reconstruct the missing texture. GAN-based inpainting models, albeit effective for general image completion tasks [58, 59, 60], are less suited for the fine-textured nature of iris images, due to training instability and limited ability to reproduce complex textures. The authors propose I3FDM, a new iris inpainting method based on denoising diffusion probabilistic models (DDPMs) [20, 21], which are known for producing high-quality, realistic images with stable training. The diffusion models work by starting with random noise and gradually denoising it step by step to generate a realistic iris image. During inpainting, the generation is conditioned on the masked iris image, meaning the known regions guide the model to fill in the missing parts. To further improve semantic consistency between the generated and original regions, the authors introduce an inverse fusion process. The model repeatedly mixes the generated image with the visible parts of the original image and then runs it through the denoising steps again. Thanks to this iterative refinement the inpainted areas align with the real parts, making the final image look more natural and better at keeping the person's identity, according to the authors' claim. The method was evaluated on the CASIA-Iris dataset. Compared to several state-of-the-art inpainting methods (including RN, AOT, MED and LaMa), I3FDM achieved superior performance in both visual quality and biometric recognition. The inpainted samples were evaluated with OSIRIS iris recognition system and the methods obtained EER of, respectively, 0.1534, 0.1095, 0.0996, 0.0707 and I3FDM gaining 0.0683.

Bhuiyan *et al*. performed iris synthesis of infant iris images - an important application,as the quantity of real samples is restricted and hardly obtainable due to ethical concerns, practical constraints, and the specialized conditions required during collection [61]. The authors managed to generate the new set of artificial infant identities having only 1920 images of 17 subjects. They trained StyleGAN2-ADA and produced a database of 5000 images. To ensure the generated images were not overly similar to real infant irises each synthetic iris sample was compared with the training dataset using the HDBIF matcher. If a Hamming distance between a synthetic and real sample was below 0.5, the synthetic one was discarded. The resulting dataset includes 1000 synthetic iris images, representing 500 new identities

Pal *et al*. addressed the challenge of poor cross-domain generalization in iris presentation attack detection (PAD) systems, which struggle when faced with unseen attack types or unfamiliar sensors [62]. To overcome this, they proposed a data augmentation technique using a convolutional autoencoder ADV-GEN, designed to generate synthetic adversarial iris images. This model inputs one iris image (bonafide or attack) and a transformation parameter vector with geometric and photometric transformations such as translation, rotation, scaling, brightness, contrast, shear, and solarization. The autoencoder is trained using a dual-loss approach: a reconstruction loss, which ensures that the output resembles a realistically transformed image, and an adversarial loss, which encourages the synthetic image to be misclassified by a pre-trained DenseNet-121 PAD classifier. Unlike traditional adversarial techniques that inject noise, ADV-GEN leverages transformation parameters to produce visually realistic yet challenging adversarial examples, enriching training diversity. Their method significantly improved PAD generalization capability, especially in cross-domain scenarios, which are critical for real-world deployment. Across experiments using benchmark datasets LivDet-Iris 2017 & 2020, the PAD classifier trained on augmented data consistently outperformed the baseline standard PAD. Correctly detected attacks at 0.1% BPCER improved by 20 to 56 percentage points..

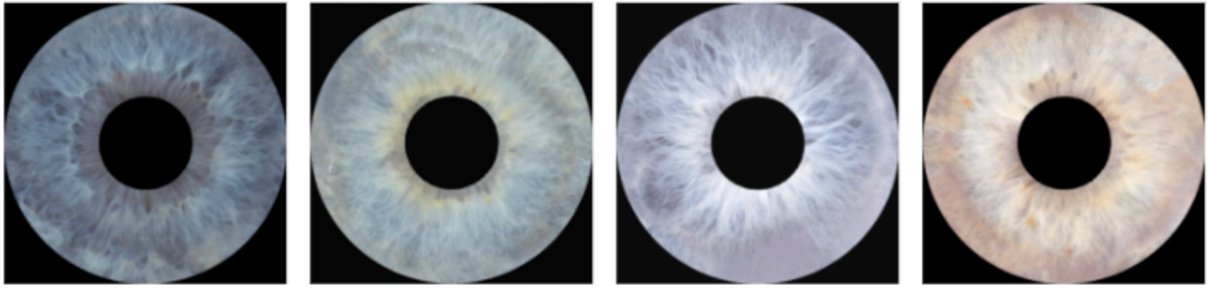

Fig. 7. Full-resolution iris textures generated with diffusion models [63].

Zhang *et al.* addressed the issue of synthetic, color iris generation utilizing the relatively new conditional diffusion models [63]. The authors trained a two-class conditional diffusion model using the Guided Diffusion framework to generate iris images conditioned on pigmentation type (blue/green or brown). Unlike previous models, this approach learns to generate full-resolution iris textures from pure noise through a gradual denoising process, guided by the desired color class, Fig. 7. The model was trained on a set of 1757 high-quality iris images. The segmented iris regions were unwrapped into polar coordinates to normalize their structure, and any resulting holes or gaps from the segmentation process were filled through imputation with information from the surrounding valid regions. Finally, the images were re-wrapped into Cartesian space and augmented through 12 rotational transformations, expanding the dataset to 21084 samples for training the diffusion model. To evaluate the biometric safety and visual quality of the generated data, the authors conducted a comprehensive analysis comparing every synthetic iris image against all real training images and confirmed that the synthetic images did not match any training samples, with a mean Hamming distance of 0.4622 and a false acceptance rate of only $3 \times 10^{-6}$ which is above the threshold used for iris matching. In terms of visual quality, the authors report a FID score equal to 0.154 at 250,000 training steps, indicating highly realistic outputs. The study demonstrates that diffusion models, though relatively new to biometric image synthesis, can generate data that is not only visually convincing but also identity-safe, marking a significant advancement in privacy-conscious biometric research.

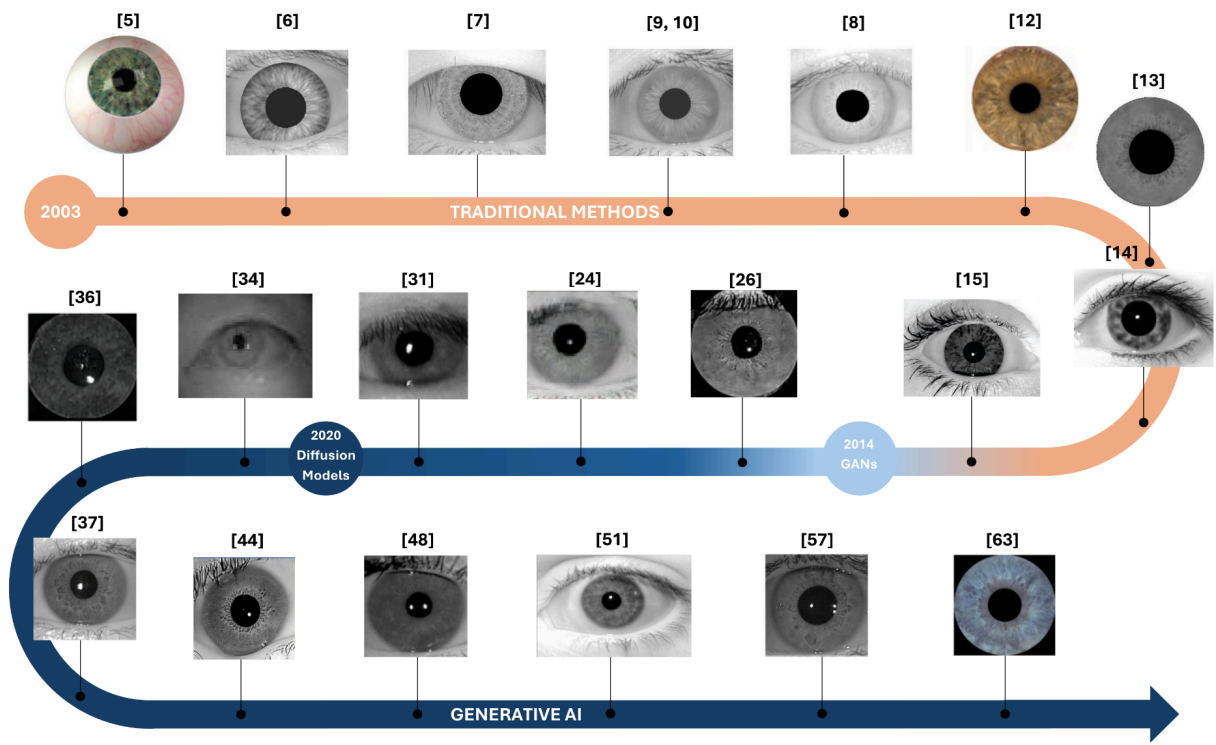

Fig. 8. Timeline of changes in the outcome of synthetic iris generation.

GANs have undergone significant advancements in the field of synthetic iris image generation, evolving from early deep convolutional architectures to more sophisticated conditional and relativistic models. Right now the state-of-the-art in synthetic iris generation are GAN variations based on StyleGANs architecture. Alongside these, autoencoder-based approaches have proven valuable for their ability to model iris deformations and data augmentation. Diffusion models have also shown promise in biometric synthesis tasks, with their capacity to produce high-quality, diverse samples [64]. This progression has been driven by the need to enhance the quality, diversity, and realism of generated iris images, which are crucial for improving biometric recognition systems and addressing challenges such as class imbalance and presentation attacks. The changes in synthetic iris generation outcomes obtained with all the described methods are shown in Fig. 8. As generative models became increasingly powerful, a new challenge emerged: identity leakage. Early GANs, due to their limited ability to create completely novel identities, often produced images that closely resembled training data. While irrelevant in early DCGAN-based models—where the generated images were still relatively low quality—it became a significant problem as realism improved.

## 4. Identity Leakage in Synthetic Datasets

With the advancement of more powerful GAN architectures, researchers noticed that some synthetic images contained biometric features strikingly similar to those of real individuals in the training set [45]. A well-documented weakness of machine learning training processes is their tendency to overfit to the training data. While commonly associated with memorizing training data rather than generalizing, in the case of generative models it manifests in a more alarming way—by embedding traces of real identities into predictions. This identity leakage phenomenon raised significant concerns related to biometric security, privacy protection, and regulatory compliance. If such models are to replace biometric data collection from real individuals, such behavior must be eliminated. The first to bring this issue to light was Tinsley *et al.,* who utilized StyleGAN2 to generate synthetic face images and subsequently evaluated them across five different face recognition systems [65]. They discovered that identity information from real training data can unintentionally appear in synthetic face images. Moreover, false match rates increased significantly for individuals whose data was used in training, when real samples were matched against generated ones. Encouraged by these findings, the same researchers turned their attention to another biometric modality: iris [45]. While some of the studies specifically sought to retain iris identity information for the purpose of intra-class data augmentation [37, 39, 48], most of them aimed to synthesize entirely new identities, not aware that identity leakage can occur. For the verification purpose the StyleGAN3 was trained on an iris dataset and the output synthetic images were examined for similarities to real training samples [45]. The outcomes of the experiment revealed that some generated irises matched training identities nearly perfect, confirming identity leakage, although only for a subset of synthetic images. By analyzing the training process at different stages, researchers observed that models trained for more iterations resulted in higher unintended identity replication. The privacy implications which arise with the identity leakage is that having a compromised model, it is possible to reconstruct the real iris images from the training set. As researchers delved deeper, they began wondering if identity leakage could be systematically traced. This led to the development of IdProv, a method designed to uncover the origins of leaked identity signals in face images [66]. Rather than just detecting leakage, IdProv analyzes the composition of synthetic faces and systematically retrieves real identities that contributed to them. By comparing cosine distances between face embeddings, extracted from images using ArcFace, the method can confidently trace back synthetic identity signals to their real counterparts. If identity leakage is present, the distance between a synthetic image and its original source is significantly lower.

Singh *et al*. introduced the SynthProv framework to analyze identity leakage in StyleGAN2's latent space [67]. They identified which specific training images are responsible for identity leakage and explained the underlying processes. The framework creates composite synthetic images through latent space interpolation and then compares them to real training images using face recognition models ArcFace and ElasticFace. By analyzing cosine distances between embeddings, SynthProv identifies which real images contribute most to synthetic faces. A key result supporting this analysis shows that the density of real identity embeddings near synthetic composites is significantly higher than near standard synthetic images, indicating a stronger identity leakage in composite outputs. To improve interpretability, the framework identifies identity-invariant directions in the GAN's latent space, which change general facial features (like pose or lighting) without affecting identity. Finally, SynthProv ranks real images based on their similarity to synthetic ones, producing a clear and interpretable mapping of leaked identity information. Unlike previous methods that only detect leakage, SynthProv reconstructs the chain of identity inheritance, offering both transparency and diagnostic insight. It occurs that identity leakage occurred consistently across models (StyleGAN2 and StyleGAN2-ADA) and matchers (ArcFace or ElasticFace), demonstrating the generalizability of the issue.

## 5. Identity leakage prevention methods

Most of the research on identity leakage has focused on face biometrics, as facial synthesis using GANs has been widely studied and recognized as a major privacy concern. However, iris biometrics, despite being one of the most secure and stable biometric modalities, has received far less attention in this context. As iris patterns are highly structured and unique, with complex textures that are difficult to synthesize, identity leakage might behave differently in iris synthesis than in face synthesis.

Once the problem became evident, researchers specializing in iris biometrics began developing methods specifically designed to mitigate leakage risks in synthetic iris images. A key breakthrough in this effort was the method described in iWarpGAN, which disentangles identity and style through two transformation pathways, ensuring that synthetic irises do not retain identity from training data [44]. This method successfully eliminates identity leakage in synthetic iris generation by using techniques such as:

- applying nonlinear identity transformations in the latent space, which shifts generated identities away from real ones, preventing direct replication,
- separating identity and style encoding, ensuring that texture variations do not reintroduce real biometric details,
- enforcing uniqueness with specialized loss functions, which penalize the model if a synthetic iris becomes too similar to a real one.

By integrating these mechanisms, iWarpGAN effectively eliminates identity leakage, producing synthetic irises that are diverse, unique, and untraceable to real identities.

The work by Kordas *et al.* focused on generating synthetic iris images in both Cartesian and polar representations to improve their usability in biometric applications [47]. While the study did not explicitly aim to prevent identity leakage, the authors reported that using StyleGAN3, no identity leakage was observed in their biometric matching tests—meaning that the synthetic irises did not falsely match real training identities at significant rates. This suggests that StyleGAN3 may inherently reduce identity leakage, but it does not include specific identity preservation mechanisms like iWarpGAN. Unlike iWarpGAN, which is dedicated solely to generating new identities, EyePreserve offers a comprehensive iris synthesis framework that enables both identity-preserving transformations and the creation of entirely new synthetic irises [48]. To safeguard against identity leakage, EyePreserve employs adversarial identity loss and triplet-based identity matching, ensuring that synthetic irises remain biometrically distinct from real individuals in the training set. Additionally, it integrates biometric verification tools like OSIRIS and HDBIF, actively testing its generated irises to confirm they do not falsely match real identities, making it one of the most secure and privacy-conscious solutions for synthetic iris generation. When generating completely new synthetic irises, EyePreserve leverages StyleGAN2 and StyleGAN3 models to produce highly realistic yet entirely fictional irises.

Although diffusion models are less prone to mode collapse, a common GAN problem, where the model generates limited or repetitive outputs, failing to reflect the full diversity of the training data, they can still memorize input patterns. A work on diffusion models by Rahimi et al. explored an identity leakage prevention method using a mixed-class sampling strategy [68]. In this approach, synthetic identities are generated by blending the conditioning vectors, which is a numerical representation of a class or identity that tells the model what to generate, of two distinct real classes, creating novel classes that are from the originals yet internally consistent. By carefully selecting mixing weights through a grid search, the method ensures that the new synthetic identities are sufficiently far from any real identity, effectively reducing the risk of identity leakage while enriching the training set with diverse, privacy-safe samples.

What started as a curious observation of overfitting in generative models has now evolved into a major privacy and security concern. The ability to generate hyper-realistic synthetic faces and irises was once thought to be a step toward safer and more ethical AI, but these findings suggest that synthetic data is not as anonymous as was once believed. As researchers work toward developing privacy-preserving generative models, one thing is clear: without proper safeguards, the very AI tools designed to protect privacy may inadvertently become a threat to it.

## 6. Summary

Over the past decade, the field of synthetic iris image generation has undergone a remarkable transformation. What began with basic Generative Adversarial Networks (GANs) has evolved into a sophisticated landscape of advanced architectures, including relativistic GANs, conditional GANs, autoencoder-based models, diffusion models and hybrid approaches that combine multiple deep learning techniques. These advancements have not only enhanced the visual realism and texture fidelity of synthetic irises but have also paved the way for greater control over identity management—a crucial aspect of biometric data synthesis.

With rising privacy and ethics-related concerns over the collection of vast, diverse, and inclusive biometric datasets, these generative AI-assisted synthesis methods can allow for an uninterrupted development of biometric methods. Applying these techniques, however, must be done with caution, as data leakage from the training sets into the network inferences has been observed across many studies. Since the entire point of synthesizing biometric data is to protect the privacy of the people, these behaviors are not acceptable and have to be mitigated. Fortunately, several data leakage alleviation strategies have been proposed to date, giving hope for a broad use of synthetic datasets, not only in iris image generation, but for biometric samples in general as well.